\documentclass[conference]{IEEEtran}
\IEEEoverridecommandlockouts
\usepackage{cite}
\usepackage{amsmath,amssymb,amsfonts}
\usepackage{algorithmic}
\usepackage{graphicx}
\usepackage{textcomp}
\usepackage{xcolor}
\usepackage{hyperref}
\usepackage{listings}
\usepackage{wrapfig}

\definecolor{codegreen}{rgb}{0,0.6,0}
\definecolor{codegray}{rgb}{0.5,0.5,0.5}
\definecolor{codepurple}{rgb}{0.58,0,0.82}
\definecolor{backcolour}{rgb}{0.94,0.94,0.94}

\lstdefinestyle{mystyle}{   
    backgroundcolor=\color{backcolour},  
    framexleftmargin=1mm,
    framexrightmargin=1mm,
    xleftmargin=1mm,
    xrightmargin=1mm,
    basicstyle=\linespread{1}\scriptsize,
    breakatwhitespace=true,         
    breaklines=true,                 
    captionpos=b,        
    keepspaces=true,                
    showspaces=false,                
    showstringspaces=false,
    showtabs=false,           
    columns=fullflexible
}

\def\BibTeX{{\rm B\kern-.05em{\sc i\kern-.025em b}\kern-.08em
    T\kern-.1667em\lower.7ex\hbox{E}\kern-.125emX}}
\begin{document}



\title{InsightGUIDE: An Opinionated AI Assistant for Guided Critical Reading of Scientific Literature\\
\thanks{
\protect\begin{wrapfigure}[3]{l}{0.9cm}
\protect\raisebox{-12.5pt}[0pt][7pt]{\protect\includegraphics[height=.9cm]{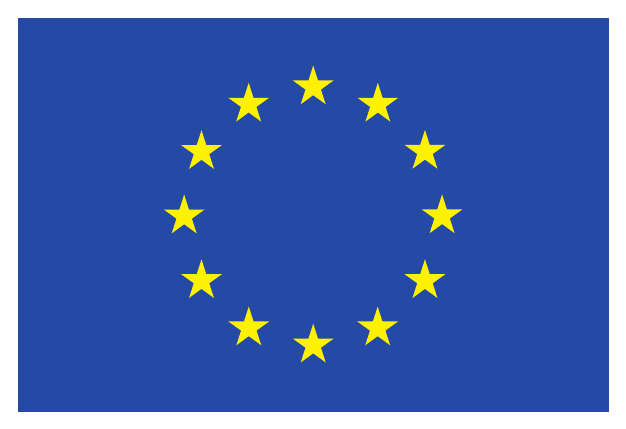}}
\protect\end{wrapfigure}%
\hspace{1mm}
\begin{minipage}[t]{\dimexpr\columnwidth-0.9cm-1mm\relax}
This work has received funding from the EU's Horizon Europe \\ framework as part of the SciLake project (GA: 101058573).
\end{minipage}
}
}

\makeatletter
\newcommand{\linebreakand}{%
  \end{@IEEEauthorhalign}
  \hfill\mbox{}\par
  \mbox{}\hfill\begin{@IEEEauthorhalign}
}
\makeatother


\author{\IEEEauthorblockN{Paris Koloveas}
\IEEEauthorblockA{\textit{Information Management Systems Institute} \\
\textit{ATHENA Research Centre}\\
Athens, Greece \\
pkoloveas@athenarc.gr -- 0000-0003-2376-089X}
\and
\IEEEauthorblockN{Serafeim Chatzopoulos}
\IEEEauthorblockA{\textit{Information Management Systems Institute} \\
\textit{ATHENA Research Centre}\\
Athens, Greece \\
schatz@athenarc.gr -- 0000-0003-1714-5225}
\linebreakand
\IEEEauthorblockN{Thanasis Vergoulis}
\IEEEauthorblockA{\textit{Information Management Systems Institute} \\
\textit{ATHENA Research Centre}\\
Athens, Greece \\
vergoulis@athenarc.gr -- 0000-0003-0555-4128}
\and
\IEEEauthorblockN{Christos Tryfonopoulos}
\IEEEauthorblockA{\textit{Department of Informatics and Telecommunications} \\
\textit{University of the Peloponnese}\\
Tripolis, Greece \\
trifon@uop.gr -- 0000-0003-0640-9088}
}


\maketitle

\begin{abstract}

The proliferation of scientific literature presents an increasingly significant challenge for researchers. While Large Language Models (LLMs) offer promise, existing tools often provide verbose summaries that risk replacing, rather than assisting, the reading of the source material. This paper introduces InsightGUIDE, a novel AI-powered tool designed to function as a reading assistant, not a replacement. Our system provides concise, structured insights that act as a ``map'' to a paper's key elements by embedding an expert's reading methodology directly into its core AI logic. We present the system's architecture, its prompt-driven methodology, and a qualitative case study comparing its output to a general-purpose LLM. The results demonstrate that InsightGUIDE produces more structured and actionable guidance, serving as a more effective tool for the modern researcher.
\end{abstract}


\section{Introduction}\label{sec:introduction}

The continuous growth of scientific literature presents a significant challenge to researchers, making it increasingly difficult to engage in the deep, critical reading necessary for true comprehension and innovation. While Large Language Models (LLMs) like ChatGPT, Claude, and Gemini, offer a potential solution, the dominant paradigms for their application, generic summarization and conversational interfaces, present significant drawbacks for scholarly use. These approaches often obscure the source material and fail to foster the critical analysis skills essential for research.

In this work, we argue for a different paradigm: an AI tool that functions as a reading assistant, not a replacement. We introduce \textbf{InsightGUIDE} (\textbf{G}uided \textbf{U}nderstanding and \textbf{I}ntelligence for \textbf{D}ocument \textbf{E}xploration), a novel system that operationalizes established expert reading methodologies to guide users through a structured, analytical process. Instead of producing verbose summaries or reactive chat responses, InsightGUIDE proactively generates a concise, structured ``map'' of a paper's key elements, presented in a dual-pane interface that keeps the source document at the center of the user's attention.

This paper makes the following contributions: 

\begin{enumerate}
    \item The design and implementation of InsightGUIDE, 
    \item A methodology for encoding expert heuristics into an LLM via a structured system prompt,  
    \item A preliminary qualitative evaluation demonstrating our system's utility, and
    \item The open-sourcing of the entire system, with the code, system prompts, and a live demonstration available for community use and evaluation\footnote{\url{https://github.com/pkoloveas/InsightGUIDE-ICTAI25}}.
\end{enumerate}

\begin{figure*}
    \centering
    \includegraphics[width=1\textwidth]{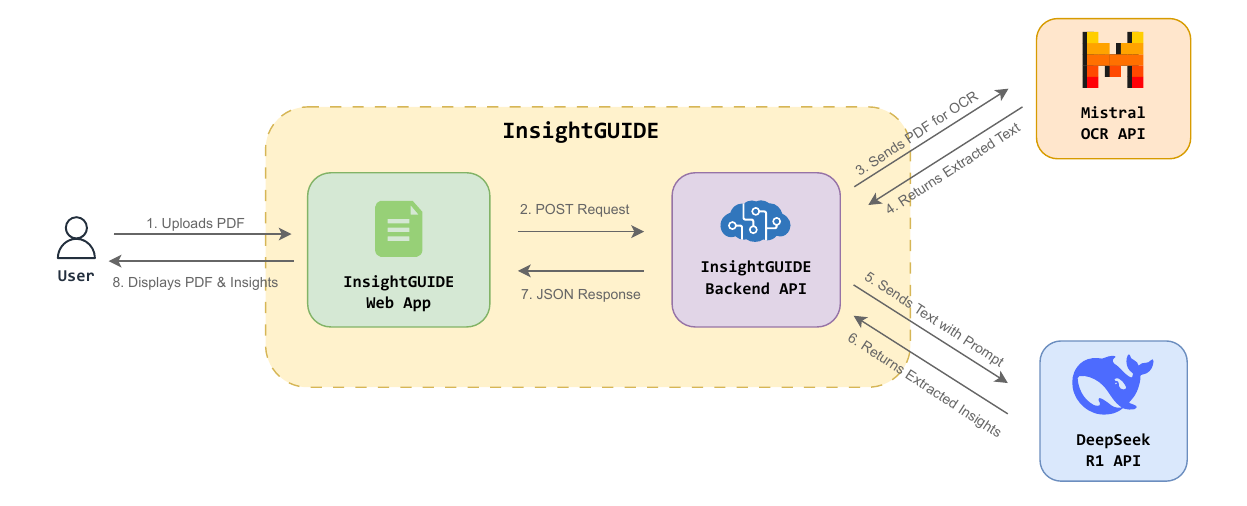}
    \caption{A High-level System Overview with the Complete User Workflow.}
    \label{fig:system-architecture}
\end{figure*}

\section{Related Work}\label{sec:related_work}







Our work is situated at the intersection of AI-driven document analysis and the pedagogy of scholarly reading. We first review the state-of-the-art in automated document synthesis to contextualize our contribution, then establish the methodological foundation for our approach.

\subsection{AI for Scholarly Document Analysis}

Automating the synthesis of scientific literature is a well-established research area, with recent surveys highlighting the rapid progress driven by LLMs~\cite{genaidiscovery, luo2025llm4sr}. However, the two most common application paradigms present critical flaws.

The first is the generic summarizer. Its primary function is to condense a source text into a shorter summary. Early approaches were primarily extractive, selecting salient sentences from source documents, for instance by using citation context~\cite{cohansumm} or sentence-BERT models~\cite{jyoti2025deep}. While effective at identifying key phrases, they lack narrative coherence. Modern abstractive systems generate new prose, but often produce verbose, unstructured text that adds to the user's cognitive load and encourages passive consumption. An evolution of this is structured summarization, where systems like BigSurvey~\cite{luistructuredsum} and CS-PAPERSUM~\cite{liu2025cs} populate predefined templates (e.g., ``methodology'', ``key takeaways''). While an improvement, their goal remains summarization within a fixed schema, not guiding the user's own analytical process.

The second dominant paradigm is the conversational interface for DocQA (Document QuestionAnswering), often marketed as a ``chat with your PDF'' service, seen in tools like  pdf.ai\footnote{\url{https://pdf.ai}}, Elicit\footnote{\url{https://elicit.com/}}, and SciSpace\footnote{\url{https://scispace.com/}}. This approach abstracts the source material behind a conversational layer, making the quality of output entirely dependent on the user's ability to formulate precise questions, a significant barrier for non-experts. This interaction model not only hides the source text but also introduces a critical vulnerability: abstractive models are highly prone to producing content unfaithful to the source document~\cite{maynez-etal-2020-faithfulness}, and the diagnosis of such factual hallucinations remains a non-trivial problem~\cite{cao-etal-2022-hallucinated}. More advanced Retrieval-Augmented Generation (RAG) systems like CORE-GPT~\cite{coregpt} and LitLLMs~\cite{litllms} mitigate this by grounding responses in retrieved documents, but their focus is typically on multi-document question-answering or literature review generation.

In contrast, InsightGUIDE proposes a third paradigm. It avoids the pitfalls of summarization by providing concise, analytical points rather than verbose prose. It avoids the risks of conversational interfaces by adopting a static, proactive approach that maps out the entire paper's structure upfront, keeping the user grounded in the source text for deep, single-document analysis.

\subsection{Foundational Methodologies for Reading Research}

The core analytical framework of InsightGUIDE is not arbitrary, but is grounded in established expert methodologies for reading scientific papers. These strategies, championed by prominent researchers like Andrew Ng in his famous Stanford lecture~\cite{ng2018reading} and S. Keshav~\cite{keshav2007read}, consistently reject a linear, cover-to-cover approach. Instead, they advocate for a multi-pass strategy where a reader first skims for high-level structure, understanding the core problem, contributions, and conclusions, before diving into the specifics of the methodology. This principle is also echoed in research guides from academic institutions such as the University of Southern California (USC)~\cite{usc_reading_research_2025}. A second key principle is the need for active critical questioning throughout the reading process, where the reader constantly assesses the validity of the claims and the suitability of the methods~\cite{keshav2007read}. InsightGUIDE is explicitly designed to automate and scaffold this process for the user.


\section{The InsightGUIDE System}\label{sec:system_architecture}

\begin{figure*}[t]
    \centering
    \includegraphics[width=1\textwidth]{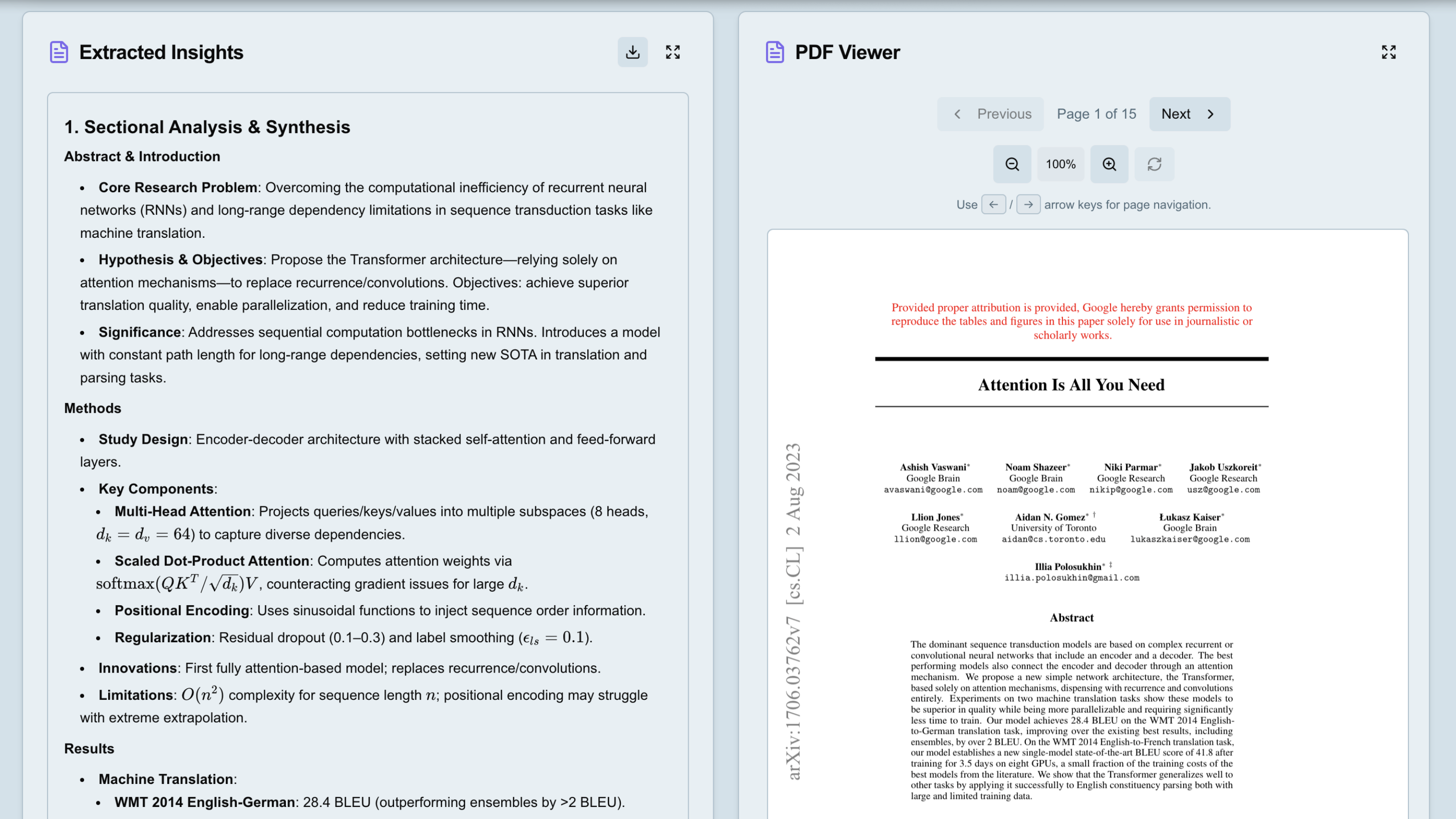}
    \caption{Snapshot of the dual-pane UI with the Extracted Insights (left) and PDF Viewer (right).}
    \label{fig:UI-loaded-insights}
\end{figure*}

\subsection{System Architecture and Workflow}


InsightGUIDE is implemented as a modern client-server web application. The system's architecture, illustrated in Figure~\ref{fig:system-architecture}, comprises a frontend client that communicates with a backend API, which in turn orchestrates calls to external AI services.

The user workflow is designed for simplicity and directness. A user first loads a PDF into the frontend client, either by local upload, from a public URL, or by selecting a pre-loaded example. When the user initiates the analysis, the frontend sends the document to the backend API. The backend then executes a two-stage AI process: first, it uses an OCR service to extract the paper's text, and second, it uses an LLM guided by our system prompt to generate the analytical insights. Finally, the structured insights are returned to the client and displayed alongside the source document.

\subsection{Frontend Client}
The client is a single-page application built with Next.js and React, designed to reinforce InsightGUIDE's philosophy of being an assistant to the researcher. Its central feature is the dual-pane user interface (Figure~\ref{fig:UI-loaded-insights}), which presents the generated insights alongside the original PDF. This design keeps the source document as the primary focus, encouraging active cross-referencing and preventing the AI-generated content from replacing the act of reading. To accommodate shifts in focus, the interface allows either pane to be maximized, enabling the user to concentrate fully on either the source text or the generated analysis as needed.

\subsection{Backend API Service}

The backend is an API service built in Python with FastAPI, chosen for its asynchronous capabilities which are essential for managing long-running calls to external AI services.  It exposes two primary RESTful API endpoints: one for the complete analysis pipeline (OCR and insight generation), and another for OCR-only text extraction. Upon receiving a file for full analysis, the service first sends it to the Mistral OCR API~\cite{mistral2025ocr} to perform high-fidelity text extraction. The extracted Markdown text is then combined with our curated system prompt and sent to an OpenAI API-compatible model. While the system currently uses DeepSeek-R1~\cite{deepseekai2025deepseekr1incentivizingreasoningcapability}, its modular design allows for any compatible model, including other proprietary or locally-hosted alternatives, to be used. The resulting structured analysis is returned to the client as a JSON response. While it serves the InsightGUIDE frontend, the API is also designed to be fully functional as a standalone service, allowing for integration into other research workflows.

\section{The Methodological Core: Engineering an Opinionated AI}\label{sec:methodology}










The novelty of InsightGUIDE lies in how it instructs the model to behave. We achieve this through a detailed, structured system prompt that translates the abstract principles of expert reading methodologies into a concrete set of instructions, forcing the LLM to move beyond summarization and act as a structured, critical guide. The design is grounded in established strategies that prioritize a deep, analytical understanding of a paper's structure and arguments. Our system prompt\footnote{\url{https://github.com/pkoloveas/InsightGUIDE-ICTAI25/blob/main/InsightGUIDE/backend-api/system_prompts.yaml}} operationalizes this theory through three key components.


\begin{table*}[t]
    \scriptsize
    \caption{Comparative Analysis of Generated Outputs}
    \renewcommand{\arraystretch}{1.4}
    \begin{center}
        
    \begin{tabular}{|l|p{6cm}|p{6cm}|}
    \hline
        \textbf{Dimension of Analysis}              & \textbf{InsightGUIDE} & \textbf{Baseline LLM (DeepSeek-R1)} \\
        \hline
        Structural Deconstruction                   & Provides distinct analysis for Abstract \& Introduction, Methods, Results, and Discussion sections. & Provides a single, monolithic summary paragraph. \\
        \hline
        Identification of Key Contribution          & ``\textbf{Transformer Architecture}: First purely attention-based sequence model, enabling parallelization.'' & ``\ldots introduces the Transformer, a novel neural
        network architecture for sequence transduction tasks such as\ldots''  \\
        \hline
        Highlighting of Methodological Limitations  & ``$O(n^2)$ \emph{Complexity}: Scalability concern for very long sequences.'' & \emph{(No specific limitations are mentioned.)} \\
        \hline
        Preemptive Critical Questions               & ``\textbf{Problem-Method Alignment}: \emph{Attention eliminates sequential computation, directly addressing RNN bottlenecks\ldots}'' & \emph{(No critical questions are posed or answered.)} \\
        \hline
        Reference to In-Paper Evidence              & ``\emph{Table 2}: Critical for comparing SOTA results and training costs.'' & \emph{(No references to specific tables or figures.)} \\
        \hline
        Actionable Guidance for the Reader          & \textbf{``For Technical Implementation}: Start with \textbf{Section 3 (Model Architecture)} $\rightarrow$ \textbf{Section 5 (Training)}\ldots'' & \emph{(No equivalent navigational guidance provided.)} \\
        \hline
        Output Format                               & Structured with headers, bullet points, and visual ``Priority Signals''. & A single, dense paragraph of continuous prose. \\
    \hline
    \end{tabular}
    \label{tab:comparative_analysis}
    \end{center}
\end{table*}

\subsubsection{Sectional Analysis \& Synthesis}

Expert reading strategies emphasize a systematic deconstruction of a paper by its core sections~\cite{ng2018reading, keshav2007read, usc_reading_research_2025}. Our prompt directly implements this by instructing the LLM to analyze the Abstract, Introduction, Methods, and Results sections individually. For each, it must extract specific elements, such as the \emph{core research problem} from the introduction, the \emph{innovative approaches} in the methodology, and the \emph{key findings} from the results. This mirrors the initial ``passes'' an expert reader makes to grasp the high-level structure and purpose of a paper before engaging in a deeper critique.

\subsubsection{Critical Evaluation \& Attention Signals}

A core tenet of effective reading is to engage in critical thinking throughout the process by constantly asking questions. This component elevates the system's output from a passive summary to an active analysis by instructing the LLM to do just that. It must identify the paper's primary \emph{Key Contributions} and, most importantly, to preemptively answer a set of \emph{Critical Questions} like ``Are the conclusions supported by data?''. This forces the model to adopt an analytical stance. Furthermore, the prompt instructs the model to use ``Priority Signals'', specific icons to flag innovative concepts, methodological limitations, or high-impact figures, respectively. This provides the user with immediate visual cues to the most salient points of the paper, aiding their own critical assessment.

\subsubsection{Reader Guidance}

Advanced reading methodologies recognize that a linear, cover-to-cover approach is often inefficient. The final component of our prompt provides direct, actionable advice that implements this non-linear strategy. It requires the generation of ``Non-Linear Navigation Tips'', which suggest efficient reading paths tailored to different user goals (e.g.,~technical replication versus a quick overview). This transforms the AI from a simple information extractor into a genuine reading guide that adapts to the user's specific objectives.

\section{A Preliminary Qualitative Evaluation}\label{sec:evaluation}

To assess the practical utility of InsightGUIDE, we conducted a preliminary qualitative evaluation. This case study was designed not as an exhaustive user study, but as a comparative analysis to highlight the fundamental differences in output between our opinionated system and a general-purpose LLM used as a summarizer\footnote{The complete evaluation outputs are available in the GitHub repository.}.

\subsection{Comparative Analysis}

Our research question for this evaluation was: ``How does the output of InsightGUIDE compare to that of the same underlying LLM using a generic prompt?''. By using the same 

\newpage
\begin{lstlisting}[caption=Summary output of Baseline LLM, label=lst:baseline]
The paper ``Attention Is All You Need'' introduces the Transformer, a novel neural
network architecture for sequence transduction tasks such as machine translation.
Unlike previous state-of-the-art models that rely on recurrent or convolutional 
layers, the Transformer is based entirely on attention mechanisms, specifically 
self-attention and multi-head attention, eliminating the need for recurrence and 
convolutions. This design allows for greater parallelization during training, leading 
to significant improvements in both computational efficiency and model perform-
ance. The Transformer achieves superior results on machine translation benchmarks, 
notably setting new state-of-the-art BLEU scores on the WMT 2014 English-
to-German (28.4) and English-to-French (41.8) tasks, while requiring less training 
time and computational resources compared to prior models. The architecture 
leverages positional encodings to maintain information about the order of sequences, 
employs residual connections and layer normalization for stability, and uses 
advanced regularization techniques. Beyond translation, the Transformer demon-
strates strong generalization to other tasks, such as English constituency parsing. 
The paper concludes that attention-based models like the Transformer offer a pro-
mising direction for future research in sequence modeling across various domains.
\end{lstlisting}

\noindent model for both our system and the baseline, we isolate the impact of the prompt itself.

To answer this, we selected the highly influential paper ``Attention Is All You Need''~\cite{vaswani2017attention} as our test case. We processed the paper using two methods:
\begin{enumerate}
    \item \textbf{InsightGUIDE:} The paper was analyzed using our system, which uses a structured prompt to instruct the DeepSeek-R1 model.
    \item \textbf{Baseline LLM:} The full text of the paper was provided to the same DeepSeek-R1 model with the generic prompt, ``Summarize the provided paper''.
\end{enumerate}

\subsection{Findings}

The results of this preliminary evaluation, as summarized in Table~\ref{tab:comparative_analysis}, indicate a clear qualitative difference between the two outputs, despite both being generated by the same underlying model. While the baseline LLM (Listing~\ref{lst:baseline}) provides a factually accurate summary of the paper's content, its output is unstructured and lacks analytical depth. It successfully answers ``What does this paper say?'' but fails to address the more critical needs of a researcher.

In contrast, InsightGUIDE's output is demonstrably more useful as a reading guide. By deconstructing the paper section by section, explicitly identifying key contributions, flagging methodological limitations, preemptively answering critical questions, and providing actionable, non-linear reading paths, it actively scaffolds the user's critical engagement with the source material. This suggests that our opinionated, methodology-driven prompt is a key factor in producing a more effective output for AI-assisted scholarly reading, transforming a generic summarizer into a structured analytical tool.

\section{Discussion and Future Work}\label{sec:duscission_future_work}

The preliminary evaluation demonstrates that our methodology of using a structured, opinionated prompt adds significant value, transforming a general-purpose LLM into a specialized tool for critical analysis. The results suggest that the primary bottleneck in AI-assisted reading is not the raw capability of the model, but the paradigm of interaction. By moving away from generic summarization and unstructured conversation towards a proactive, scaffolded approach, InsightGUIDE provides a more effective framework for researchers. However, it is important to acknowledge the limitations of the current system and consider avenues for future work.

\subsection{Limitations}

The quality of InsightGUIDE's output is inherently dependent on the quality of its upstream dependencies: the OCR service (Mistral OCR) and the LLM (DeepSeek-R1). Errors or inaccuracies in the text extraction or the model's generation will propagate to the final output. While our structured prompt mitigates the risk of unconstrained hallucination, it does not eliminate it entirely.

Furthermore, the core ``opinion'' of the system is currently static. The system prompt is designed as a one-size-fits-all solution based on general principles of reading empirical research papers. This approach may not be optimal for all document types, such as literature reviews, theoretical papers, or survey articles, which have different structures and goals.

\subsection{Future Directions}

The limitations of the current system point toward several clear directions for future research and development. To address the static nature of the prompt, we plan to introduce a selection of predefined ``reading profiles'' tailored to different document types (e.g., ``Empirical Study'', ``Literature Review'', ``Theoretical Paper'').

Another next step for the continuation of this work would be to move beyond our preliminary case study and conduct a large-scale user study across several academic fields. Such a study would aim to formally quantify the impact of our tool on key metrics, such as the time required to understand a paper and the depth of a researcher's comprehension and critical analysis. Finally, a long-term goal is to expand the system's capabilities from single-document analysis to multi-document comparative analysis, enabling users to synthesize insights across a body of literature for tasks such as creating related work sections.

\section{Conclusion}\label{sec:conclusion}

This paper introduced InsightGUIDE, an AI-powered tool designed to address the shortcomings of generic LLM summarizers and conversational agents for scholarly reading. We have argued that by operationalizing expert reading methodologies within a structured system prompt, it is possible to create an AI assistant that guides critical engagement rather than replacing it. Our core contribution is not just the tool itself, but the methodology for creating such an ``opinionated'' AI tool that functions as a scaffold for the user's own analytical process. By keeping the user and the source document at the center of the interaction, this paradigm offers a promising direction for the future of human-AI collaboration in research, fostering deeper understanding and more efficient analysis.


\bibliographystyle{IEEEtran}
\bibliography{thebib}

\end{document}